\documentclass[conference]{IEEEtran}
\IEEEoverridecommandlockouts

\usepackage{cite}
\usepackage{amsmath,amssymb,amsfonts}
\usepackage{algorithmic}
\usepackage{textcomp}
\usepackage{xcolor}
\usepackage{booktabs}    
\usepackage{subcaption}
\usepackage{multirow}    
\usepackage{graphicx}    
\usepackage[table]{xcolor}
\definecolor{best}{RGB}{198,239,206}    
\definecolor{second}{RGB}{255,235,156}
\def\BibTeX{{\rm B\kern-.05em{\sc i\kern-.025em b}\kern-.08em
    T\kern-.1667em\lower.7ex\hbox{E}\kern-.125emX}}
\usepackage{soul}
\usepackage{amsthm}
\usepackage{mathrsfs}
\theoremstyle{plain}
\usepackage{url}

\theoremstyle{remark}

\begin{document}

\title{Task-Aware Federated Fine-Tuning for MoE-based Large Language Models\\
}

\author{
\IEEEauthorblockN{
Tingqi Wang,
Hongyu Ke,
Haoxin Wang,
Rafal Angryk,
and Zhipeng Cai\textsuperscript{*}
}
\IEEEauthorblockA{
\textit{Department of Computer Science, Georgia State University} \\
Atlanta, GA, United States \\
\{twang33, hke3, haoxinwang, angryk, zcai\}@gsu.edu
}
}

\maketitle

\begingroup
\renewcommand{\thefootnote}{\fnsymbol{footnote}}
\footnotetext[1]{To whom correspondence should be addressed}
\endgroup

\begin{abstract}
Mixture-of-Experts (MoE) has become a widely adopted architecture for Large Language Models (LLMs), as it improves model capacity while limiting computational overhead through sparse expert activation.
This property makes MoE-based LLMs particularly attractive for resource-constrained distributed environments.
However, federated fine-tuning of MoE-based LLMs remains challenging under heterogeneous client data. 
Since clients often correspond to different task preferences, directly aggregating their local updates may weaken expert specialization and introduce conflicting update directions on shared experts.
To address these challenges, we propose FedTAR, a task-aware federated fine-tuning method for MoE-based LLMs. FedTAR establishes the association between local updates and task preference via routing outputs.
Specifically, we apply Singular Value Decomposition (SVD) to both routing features and local updates to extract low-dimensional task coordinates and update directions. 
Based on the task coordinates, FedTAR performs intra-cluster aggregation among clients with similar task preferences and inter-cluster aggregation across different task groups. 
The aggregated update is then reconstructed through the learned task-to-update mapping, ensuring that the final update remains aligned with task-specific optimization directions. 
In this way, FedTAR preserves expert specialization and mitigates destructive interference among heterogeneous clients. 
We evaluate FedTAR on four benchmark tasks under different non-IID settings. Experimental results demonstrate that FedTAR consistently outperforms strong federated fine-tuning baselines and achieves state-of-the-art performance. The code is available at \url{https://github.com/Tingqi0708/FedTAR}.
\end{abstract}


\section{Introduction}
Mixture-of-Experts (MoE), a popular architecture for Large Language Models (LLMs), has attracted significant attention for its ability to balance model capability and computational overhead \cite{mu2025comprehensive, zhang2025mixture}. 
By employing conditional computation, MoE selectively activates only a subset of experts for each token \cite{zhou2022mixture}. 
Such sparse activation enables MoE-based LLMs to scale model capacity without proportionally increasing inference cost, making them well-suited for resource-constrained distributed environments \cite{zhuang2024litemoe, wang2025d2moe, ke2024carboncp}. 
In practice, adapting LLMs to various domain-specific tasks often requires fine-tuning on limited and dispersed private data owned by multiple clients \cite{wang2025federated,wu2024fedbiot}. 
Federated fine-tuning provides a promising solution by enabling collaborative adaptation without direct data sharing. However, the combination of MoE architectures and heterogeneous client data introduces new challenges that have not been systematically investigated in existing works: \textbf{(i) weakened expert specialization}, and \textbf{(ii) conflicting updates on shared experts}.

\begin{figure}[t]
\centering
\begin{subfigure}{0.48\columnwidth}
    \centering
    \includegraphics[width=\linewidth]{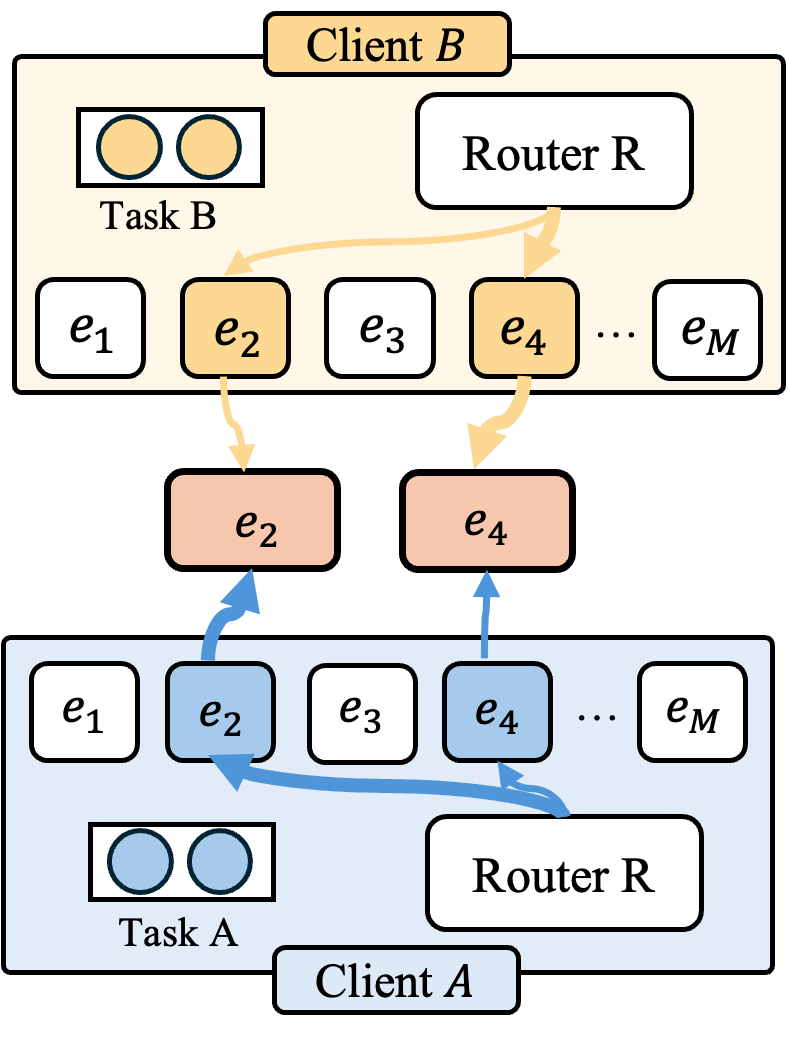}
    \caption{Expert specialization weaken due to updates averaged out.}
\end{subfigure}
\hfill
\begin{subfigure}{0.487\columnwidth}
    \centering
    \includegraphics[width=\linewidth]{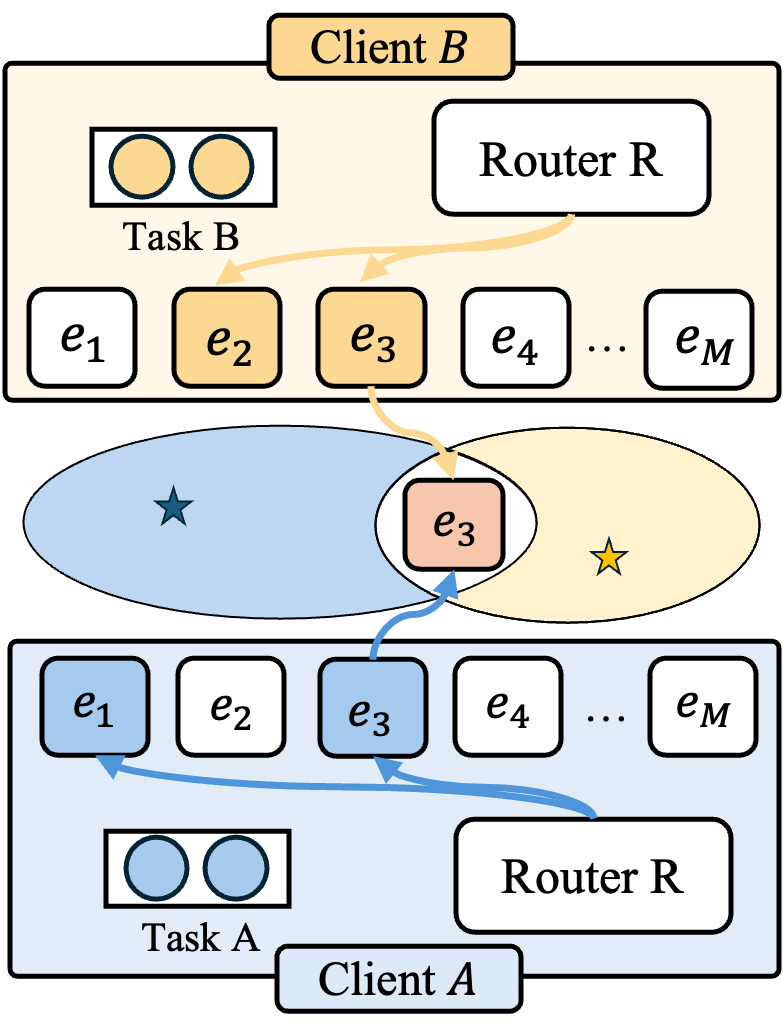}
    \caption{Share expert convergence conflicts in different tasks.}
\end{subfigure}
\caption{Illustration of challenges in Federated MoE-based LLMs Fine-Tuning.}
\label{fig:moe_fl_illustration}
\end{figure}

Specifically, clients with different task preferences may cause different update magnitudes \cite{bai2025understanding, ke2025mambev}. 
During the aggregation, these task-specific differences are averaged indiscriminately, which may weaken expert specialization across tasks. As shown in Fig. \ref{fig:moe_fl_illustration}(a), a client specializing in task A applies larger updates to expert  $e_2$ than to expert $e_4$, while a client focusing on task B exhibits the opposite pattern. 
After federated fine-tuning aggregation, the updates to $e_2$ and $e_4$ are merged without considering their task-specific roles. 
As a result, the global model may become suboptimal for both tasks A and B. 
In addition, the same expert may be activated by clients with different task preferences and thus receive updates with inconsistent optimization directions \cite{wang2025hmoe,yu2026drift,zhang2025mixture}. As illustrated in \ref{fig:moe_fl_illustration}(b), expert $e_3$ is shared by both tasks A and B, but its convergence directions under the two tasks conflict with each other. Such conflicts can cause destructive interference in global model updates and further degrade federated fine-tuning performance.

The aforementioned challenges raise an essential question: \textit{how can client updates be aligned with their underlying task intentions in federated MoE fine-tuning?} 
Since task preferences are not directly observable, it is difficult to determine which clients should be aggregated together and how their updates should be combined. 
This motivates us to introduce an intermediate representation that captures the underlying task structure. 
In MoE models, routing outputs naturally reflect expert utilization patterns of each client, and thus implicitly encode the underlying task representations. 
Based on this observation, we propose FedTAR, a novel Task-awARe Federated fine-tuning method for MoE-based LLMs. 
FedTAR leverages routing outputs to establish a mapping between the task space and the update space, enabling task-aware aggregation and reconstruction of federated updates. 
To the best of our knowledge, this is the first work focusing on federated fine-tuning for MoE-based LLMs. 
Specifically, FedTAR first collects routing statistics during local fine-tuning, including average expert utilization, expert co-activation relationships, and routing confidence for each client. 
Then, Singular Value Decomposition (SVD) is applied to both routing features and local updates to extract task coordinates and dominant update directions. 
Based on the task coordinates, FedTAR performs task-aware aggregation within and across client clusters. 
Finally, the learned mapping between task and update spaces is used to reconstruct the aggregated update, ensuring that the global update remains aligned with task-specific directions. 
In this way, FedTAR preserves expert specialization while mitigating destructive interference among heterogeneous clients. 
We conduct comprehensive experiments under different data heterogeneity levels, tasks, and hyper-parameter settings \cite{xu2025dp}. The results show that FedTAR achieves state-of-the-art (\textit{SOTA}) performance across multiple tasks and settings. 
Our contributions are summarized as follows:

\begin{itemize}
    \item To the best of our knowledge, this is the first work to study federated fine-tuning for MoE-based LLMs. We identify two unique challenges: degraded expert specialization and conflicting updates on shared experts.    
    \item We propose \textbf{FedTAR}, a task-aware federated fine-tuning method for MoE-based LLMs. FedTAR uses routing statistics to infer task representations, performs task-aware clustering and aggregation, and reconstructs global updates through a learned task-update mapping to preserve expert specialization and mitigate conflicts.
    \item We conduct extensive experiments under various data heterogeneity levels, tasks, and hyper-parameter settings. The results demonstrate the effectiveness of FedTAR and its state-of-the-art performance in federated fine-tuning.
    
\end{itemize}
\section{Related Work}
In the era of LLMs, federated learning (FL) has been widely adopted for on-device fine-tuning without direct data access.
The vanilla FL approach, FedAvg \cite{mcmahan2017communication}, averages model updates from participating clients.
However, FedAvg often underperforms on non-independent and identically distributed (Non-IID) data.
To tackle this issue, Li \textit{et al.} proposed FedProx \cite{li2020federated}, which improves training stability by adding a proximal term to local training.
SCAFFOLD mitigates client drift by employing control variates \cite{karimireddy2020scaffold}.
More broadly, alleviating performance degradation caused by Non-IID data in FL has been extensively explored \cite{ye2023heterogeneous}.
Over the years, these methods have been widely applied to various domains \cite{li2025challenges,karunamurthy2025optimal}.

Extending FL to LLMs faces additional challenges due to the massive model size and communication overhead.
To address this issue, parameter-efficient fine-tuning (PEFT) methods, such as low-rank adaptation (LoRA), have been proposed to reduce the number of trainable parameters and computational costs \cite{lin2024data}.
Recent works integrate PEFT with FL by restricting local updates to low-rank adapters and aggregating them across clients \cite{wang2025federated,gao2025federated}.
These approaches significantly improve the communication efficiency and scalability of LLM fine-tuning.
As one of the earliest approaches integrating LoRA into FL, FFA-LoRA adopts a simple yet effective strategy by freezing the LoRA $A$ matrices and only training and aggregating the LoRA $B$ matrices across clients, thereby reducing communication overhead \cite{sun2024improving}.
It also considers aggregation interference and improves robustness under differential privacy (DP) noise, enhancing training stability.
Building upon this, FedEx-LoRA further improves aggregation by introducing more flexible parameter sharing mechanisms to better capture client-specific characteristics \cite{singhal2025fedex}.
There are also works that primarily focus on improving efficiency, such as FedBiOT \cite{wu2024fedbiot} and FedDiAL \cite{yan2025feddial}.
More recently, FedSVD leverages the low-rank structure of client updates and performs aggregation in a shared subspace via SVD, which partially alleviates update inconsistency \cite{lee2025fedsvd}.

With the growing adoption of MoE architectures in LLMs, recent works have explored their integration with FL \cite{hu2025fft}.
However, these approaches mainly incorporate MoE into existing fine-tuning pipelines, rather than rethinking the federated optimization process for MoE-based models.
Consequently, they fail to fully exploit the routing mechanisms that naturally encode task-specific structures and expert specialization.
Ignoring such structural information may lead to the aggregation of incompatible expert updates across clients, especially under heterogeneous data distributions, resulting in suboptimal global models.

\section{Preliminary}
\subsection{MoE-Based LLM.}
In an MoE-based LLM, each layer contains a set of experts and a router \cite{xue2024openmoe}.
Each expert $f^{(l)}_e$ is a parameterized feed-forward network.
Given an input $x$, the router assigns the input to a subset of experts, and only the selected experts are activated.
The output of the $l$-th layer is:
\begin{equation}
h^{(l)}(x)=\sum_{e\in \mathcal{E}^{(l)}} g^{(l)}_e(x) f^{(l)}_e(x),
\quad \|g^{(l)}(x)\|_0 \le k,
\end{equation}
where $\mathcal{E}^{(l)}$ denotes the set of experts in layer $l$, $g^{(l)}_e(x)$ is the routing weight of expert $e$, and $k$ is the number of activated experts.

Compared with dense LLMs, MoE achieves a better trade-off between model capacity and computational cost.
In a dense model, all parameters are activated for each input.
In contrast, an MoE model maintains a larger set of parameters $\{W_e\}_{e\in\mathcal{E}}$, but only a subset of experts is activated for each input.
As a result, MoE enables higher model capacity without a proportional increase in computational cost.
 
\subsection{Federated Fine-Tuning}
Consider a FL system with $N$ clients $\mathcal{C}=\{\mathcal{C}_1, \mathcal{C}_2, ..., \mathcal{C}_N\}$ and a Server $\mathcal{S}$.
Each client $\mathcal{C}_i$ has a private dataset $\mathcal{D}_i$ with $n_i$ training samples $\{(x_{ij},y_{ij})\}_{j=1}^{n_i}.$.
All clients can either deploy the same MoE-based LLM locally or remotely host it.
Our objective is to fine-tune a global MoE-based LLM over these $N$ clients:
\begin{equation}
\Delta \mathbf{W}^* = \arg\min_{\Delta \mathbf{W}} \sum_{i=1}^{N} p_i \mathcal{L}_i(\mathbf{W}_0 + \Delta \mathbf{W}),
\end{equation}
where $p_i$ denotes the weight of client $i$, and $\mathcal{L}_i(\cdot)$ is the local empirical risk defined over dataset $\mathcal{D}_i$.
It should be note that our method is suitable for full fine-tuning and LoRA.
Due to the large size of foundation model, we use LoRA instead of training all parameters.
\begin{figure*}[t]
\centering
\includegraphics[width=1\linewidth]{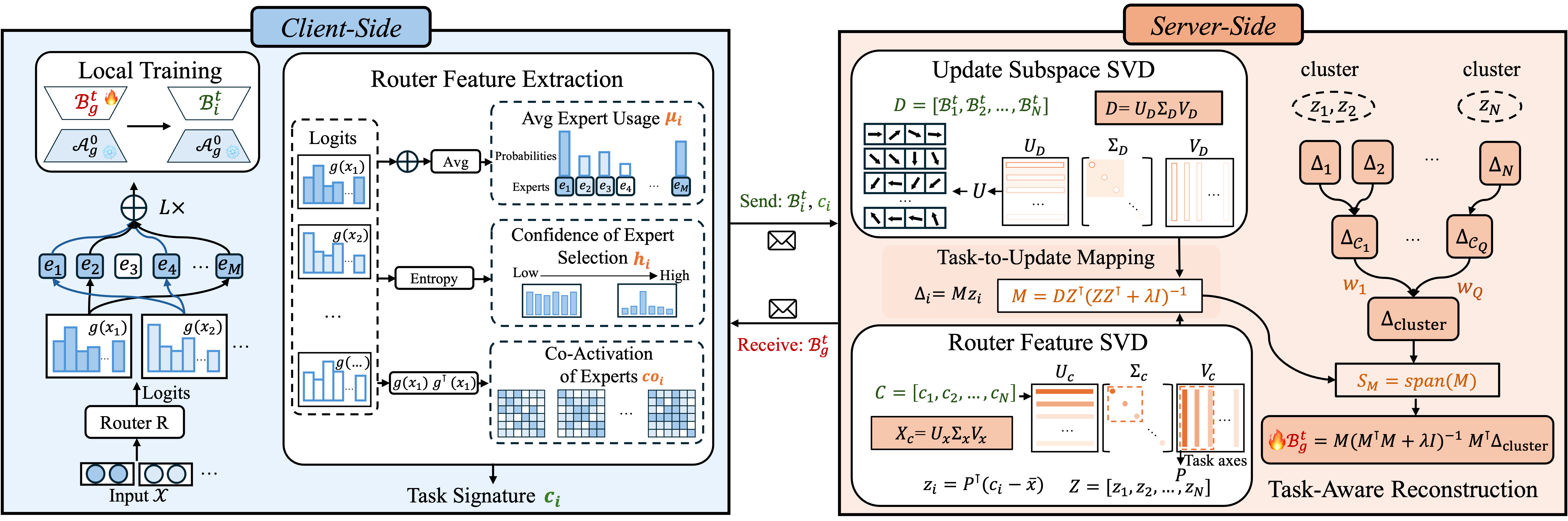}
\caption{
Overview of FedTAR. Client Side: Each client performs local LoRA fine-tuning and extracts routing features along with local updates. Server Side: The server clusters clients based on the task coordinates, aggregates updates within clusters, and reconstructs a global update via task-aligned subspace projection.
}
\label{fig:framework}
\end{figure*}

\section{Methodology}
\subsection{Overview}
We present FedTAR, a task-aware federated fine-tuning framework for MoE-based LLMs. Our design is motivated by two major challenges in federated fine-tuning of MoE-based LLMs: (i) weakened expert specialization, and (ii) incompatible update directions on shared experts caused by task heterogeneity. To tackle these issues, FedTAR captures the relationship between task preferences and model updates via local routing behaviors, thereby enabling task-aware fine-tuning of the global model.

The overview of our method is shown in  Fig. \ref{fig:framework}. On the client side, after local fine-tuning, the router calculates logits on a local held-out dataset, which are used to compute routing statistics such as mean expert usage and expert co-activation. 
The fine-tuning updates and routing features are then uploaded to the server. On the server side, FedTAR separately decomposes the local updates and routing features to extract the primary update directions and task coordinates, and then learns the mapping between them. 
Subsequently, client updates with similar task coordinates are hierarchically aggregated through clustering. Finally, guided by the learned mapping, the aggregated result is reconstructed in the task-aligned subspace to obtain the final global update. In the following subsections, we detail our design.

\subsection{Local Fine-Tuning}
All clients start with a frozen pretrained MoE model $\mathbf{W}_0 \in \mathbb{R}^{d \times k}$ deployed locally or hosted in the cloud.
Before the first communication round, the server broadcasts the global LoRA parameters $\mathcal{B}_g^{0}={\{ \mathcal{B}_{k}^{g}\}_{k=1}^{K}}$ and $\mathcal{A}_g^{0}={\{ \mathcal{A}_{k}^{g}\}_{k=1}^{K}}$ \cite{hu2022lora}:
\begin{equation}
\begin{aligned}
    & \Delta \mathbf{W}_g^{0} = \{ \mathcal{B}_{k}^{g} \mathcal{A}_k^{g} \}_{k=1}^{K}, \\
    & \mathcal{B}_{k}^{g} \in \mathbb{R}^{d_{in} \times r}, \quad
    \mathcal{A}_{k}^{g} \in \mathbb{R}^{r \times d_{out}},
\end{aligned}
\end{equation}
where $K$ is the number of experts.
$r$ is the rank of $\mathcal{B}_g^{0}\mathcal{A}_g^{0}$.
$d_{in}$ and $d_{out}$ are the input and output dimensions, respectively.
$r<<\text{min}(d_{in},d_{out})$.
$\mathcal{B}_g^{0}$ is initialized to zero, $\mathcal{A}_g^{0}$ uses random Gaussian initialization.
Following the vanilla FL LoRA setting, we freeze $\mathcal{A}_k^{g}$ and only aggregate $\mathcal{B}_g^{0}$,
enforcing a shared subspace across clients to mitigate severe update conflicts while improving communication efficiency \cite{sun2024improving}.

In $t$-th communication round, each client $\mathcal{C}_i$ fine-tunes locally on its dataset $\mathcal{D}_i$ in multiple steps. 
It optimizes the following objective:
\begin{equation}
  \mathcal{B}^t_i = \arg\min_{\mathcal{B}} \ \mathcal{L}_i\!\left(\mathbf{W}_0 + \Delta \mathbf{W}(\mathcal{B})\right).
\end{equation}
After local training, each client obtains $\mathcal{B}_i^{t}$, which captures local adaptations and is sent to the server.

\subsection{Router Feature Extraction}
In MoE models, the router selects a subset of experts for each input $x$ to process the task.
Consequently, routing behavior varies across different data distributions.
Since different tasks require distinct expert capabilities, routing decisions implicitly encode underlying task preferences, which can be leveraged to represent client-specific tasks.

Specifically, the average routing probability of each expert is a highly indicative feature.
In MoE models, the load balancing loss calculates $P_k$ for expert $k$, which denotes the average probability of assigning expert $k$ to all tokens in a batch, computed based on the routing probabilities before top-$k$ selection \cite{harvey2025optimizing, ke2026deformba}.

Inspired by this, we extend $P_k$ from the batch level to the client level.
The average routing probability $\mu_i$ for client $\mathcal{C}i$ is:
\begin{equation}
\mu_i = \mathbb{E}_{x \sim \mathcal{D}_i} [ g(x) ],
\end{equation}
where $g(x) \in \mathbb{R}^{K}$ denotes the routing probability distribution over $K$ experts for input $x$.
$\mu_i$ represents the probability that each expert is used on the current client.

Since each task may require collaboration among experts with different capabilities, we further consider expert co-activation by computing their joint usage patterns:
\begin{equation}
{co}_i = \mathbb{E}_{x \sim \mathcal{D}_i} [ g(x) g(x)^\top ].
\end{equation}
The $(p,q)$-th element of $g(x) g(x)^\top$ is given by $g_p(x) g_q(x)$, which measures the joint activation strength between experts $p$ and $q$.
Therefore, ${co}_i$ encodes the task-specific interaction structure among experts for client $\mathcal{C}_i$.

In addition, to improve robustness, we quantify the confidence of expert selection for each client by computing the entropy of the routing distribution:
\begin{equation}
h_i = \mathbb{E}_{x \sim \mathcal{D}i} \left[ - \sum_{k=1}^{K} g_k(x) \log g_k(x) \right].
\end{equation}
$h_i$ distinguishes between well-defined and ambiguous tasks, providing valuable information for aggregation: clients with low-entropy routing patterns (\textit{i.e.}, precise expert selection) tend to be more reliable.

$\mu_i$, $co_i$, and $h_i$ collectively provide a comprehensive description of each client's routing behavior.
These features capture the client's expert selection preferences, interaction structure, and selection confidence, respectively.
Then, we construct a task representation by concatenating and $\ell_2$-normalizing these features to obtain a unified and scale-invariant representation:

\begin{equation}
    c_i = \left[ \mu_i ; \mathrm{vec}(co_i) ; h_i \right], \quad \|c_i\|_2 = 1,
\end{equation}
where $\mathrm{vec}(\cdot)$ denotes matrix vectorization.
$c_i$ is sent once if the router is frozen, and every round if the router is trainable.
\subsection{Task–Update Subspace Alignment}
For each client $\mathcal{C}_i$, the server receives its update $\mathcal{B}^t_i$ and routing feature $c_i$.
Although routing features and model updates reside in different spaces, they are governed by the same underlying task variations and thus admit a shared low-dimensional structure. 
This structure enables alignment between task representations and update directions via joint subspace analysis \cite{yuan2025moore}.

To achieve this goal, we first stack all clients to form the update matrix: $\Delta = [\mathcal{B}^t_1, \mathcal{B}^t_2, \dots, \mathcal{B}^t_N]$.
Next, the SVD decomposes $\Delta$ to obtain the dominant structure of updates:
\begin{equation}
    \Delta = \mathbf{U}_\Delta \mathbf{\Sigma}_\Delta \mathbf{V}_\Delta^\top,
\end{equation}
the top-$r$ components is retained as a low-dimensional update subspace: $
    \Delta_r = \mathbf{U}_\Delta^{(r)} \mathbf{\Sigma}_\Delta^{(r)} \mathbf{V}_\Delta^{(r)\top},
$
where $\mathbf{U}_\Delta^{(r)}$ defines the principal directions of the update space, providing how to reconstruct updates from low-dimensional coordinates.
Similarly, task matrix, stacked routing features $C = [c_1, c_2, \dots, c_N]$,  can be decomposed into reorganization of task axes:
\begin{equation}
    C = \mathbf{U}_C \mathbf{\Sigma}_C \mathbf{V}_C^\top.
\end{equation}
The top-$r$ components is the low-dimensional task subspace:
$
    C_r = \mathbf{U}_C^{(r)} \mathbf{\Sigma}_C^{(r)} \mathbf{V}_C^{(r)\top},
$
where $\mathbf{U}_C^{(r)}$ captures the principal task directions, and 
$\mathbf{V}_C^{(r)}$ provides the corresponding low-dimensional task intentions for each client.
To obtain explicit task coordinates, we center the routing features as $\bar{c} = \frac{1}{N} \sum_{i=1}^N c_i, \tilde{c}_i = c_i - \bar{c}.$
Let $\mathbf{P} = \mathbf{U}_C^{(r)}$ denote the top-$r$ task subspace. 
Each client is then represented by its task coordinate, such as  $z_i = \mathbf{P}^\top \tilde{c}_i.$ It provides a geometric interpretation of the task representation:
routing feature is projected onto a shared low-dimensional task subspace,
and its differences between the various clients correspond to underlying task differences.

Then, we approximate the relationship between task representations and update directions within the learned low-dimensional subspaces using a parametric form.
Specifically, the update coordinate $\alpha_i$ is approximated from the task coordinate $z_i$ as $\alpha_i \approx \mathbf{R} z_i$,
which can be rewritten as:
\begin{equation}
    \mathcal{B}^t_i \approx \mathbf{U}_\Delta^{(r)} \mathbf{R} z_i = \mathbf{M} z_i.
\end{equation}
This formulation provides a low-dimensional approximation that links task coordinates to update directions.
With both routing features and updates projected onto their principal subspaces,
their dominant variations can be effectively aligned.
While more expressive mappings are possible, we adopt a regularized least-squares solution \cite{nguyen2024least} for stability and robustness:
\begin{equation}
    \mathbf{M} = \arg\min_{\mathbf{M}} \|\Delta - \mathbf{M} Z\|_F^2 + \lambda \|\mathbf{M}\|_F^2,
\end{equation}
which admits the closed-form solution: $
    \mathbf{M} = \Delta Z^\top (Z Z^\top + \lambda \mathbf{I})^{-1},
$
where $\lambda > 0$ is a regularization parameter that improves numerical stability and prevents overfitting.
Accordingly, for any given update $\Delta'$, we project it onto the task-aligned subspace to ensure consistency with the learned task–update structure:
\begin{equation}
    \Delta =
\mathbf{M}
(\mathbf{M}^\top \mathbf{M} + \lambda I)^{-1}
\mathbf{M}^\top \Delta'.
\end{equation}
The above calculation process is provided in the appendix.

\subsection{Task-Aware Aggregation and Reconstruction}
In server aggregation, task heterogeneity across clients significantly affects the global update.
In our method, we perform intra-cluster and inter-cluster aggregation based on task coordinates, ensuring that clients with similar tasks are aggregated together.
Then, the aggregated update is reconstructed through the mapping $\mathbf{M}$ to maintain consistency with the task-aligned subspace, thereby mitigating the negative impact of task heterogeneity.

We first cluster the task coordinates ${z_i}$ into $Q$ groups $\{\mathcal{G}_q\}_{q=1}^{Q}$, where each cluster represents a group of clients with similar task characteristics.
Within each cluster, the cluster-level update is obtained by averaging client updates: $\Delta_{c_\mathrm{q}} = \frac{1}{|\mathcal{G}_\mathrm{q}|} \sum_{i \in \mathcal{G}_\mathrm{q}} \mathcal{B}^r_i.$
Then, inter-cluster aggregation weights are assigned based on each cluster's proximity to the global task center.
Let the global task center be $z_{\text{center}} = \frac{1}{N} \sum_{i=1}^N z_i$.
The weight of each cluster is:
\begin{equation}
w_\mathrm{q} = \frac{\exp\left(-\tau | z_\mathrm{q} - z_{\text{center}} |^2 \right)}{\sum_j \exp\left(-\tau | z_\mathrm{j} - z_{\text{center}} |^2 \right)},
\end{equation}
where $z_\mathrm{q} = \frac{1}{|\mathcal{G}_\mathrm{q}|} \sum_{i \in \mathcal{G}_\mathrm{q}} z_i$ denotes the cluster-level task representation.
Therefore, the aggregated update is given by:
\begin{equation}
\Delta_{\text{Cluster}} = \sum_{q=1}^Q w_\mathrm{q} \Delta_{c_\mathrm{q}}.
\end{equation}

Although $\Delta_{\text{Cluster}}$ integrates information from different clients, it may not strictly follow the learned task--update relationship due to the aggregation of heterogeneous updates, which may introduce components outside the task-aligned subspace.
To enforce structural consistency, we reconstruct $\Delta_{\text{Cluster}}$ by projecting it onto the subspace spanned by the learned mapping $\mathbf{M}$.
Let $\mathcal{S}_M = \text{span}(\mathbf{M})$ denote the task-aligned update subspace.
The reconstructed update is given by:
\begin{equation}
\mathcal{B}_g^{t} = \mathbf{M} (\mathbf{M}^\top\mathbf{M} + \lambda \mathbf{I})^{-1} \mathbf{M}^\top \Delta_{\text{Cluster}},
\end{equation}
which is the closed-form solution of a regularized least-squares projection onto $\mathcal{S}_M$.
This projection enforces the final update to lie within the task-consistent subspace, removing components that are not aligned with the learned task--update structure.
$\mathcal{B}_g^{t}$ is the task-aware update at the current round, which is sent back to clients and applied to the global model via $\mathbf{W}_g \leftarrow \mathbf{W}_g + \mathcal{B}_g^{t}\mathcal{A}_g^{0}$.

\subsection{Privacy}
The local updates $\mathcal{B}_i^{t}$ are transmitted under differential privacy protection. 
In addition, our method requires each client to share a routing representation.
Specifically, each client computes routing statistics as $x_i = [\mu_i; \mathrm{vec}(C_i); h_i]$, where $\mu_i \in \mathbb{R}^K$ is the mean expert usage, $C_i \in \mathbb{R}^{K \times K}$ is the co-activation matrix, and $h_i$ is the entropy term.
These quantities are aggregated over local data and do not expose individual samples, but they may still reveal coarse information about the underlying data distribution.
To provide formal privacy guarantees, we apply differential privacy to the routing features via the Gaussian mechanism.
Each client first clips its feature vector to bound the sensitivity:
\begin{equation}
\bar{x}_i = x_i \cdot \min\left(1, \frac{S}{\|x_i\|_2}\right),
\end{equation}
and then adds the noise:
\begin{equation}
\tilde{x}_i = \bar{x}_i + \eta_i, \quad
\eta_i \sim \mathcal{N}(0, \sigma^2 S^2 I).
\end{equation}
This ensures that the $\ell_2$-sensitivity is bounded by $2S$, and the mechanism satisfies $(\varepsilon,\delta)$-differential privacy with $\sigma = \mathcal{O}\left(\frac{\sqrt{\log(1/\delta)}}{\varepsilon}\right)$.
Since dimensionality reduction, such as SVD, is a post-processing step, the privacy guarantee is preserved for the released representation.

\subsection{Communication Cost}
The task representation $c_i$ has dimension $\mathcal{O}(K^2)$ due to the co-activation matrix.
In each communication round, a client transmits: (i) a model update $\Delta w_i \in \mathbb{R}^d$, and (ii) the task representation $c_i$.
Thus, the communication cost per client is $\mathcal{C}_{\mathrm{ours}} = \mathcal{O}(d + K^2)$, compared with $\mathcal{O}(d)$ in standard FL.
Since $K^2 \ll d$ in practice (e.g., tens of experts versus millions of parameters), the additional overhead is small.
Moreover, $c_i$ is transmitted only once when the router is frozen, and updated every round only when the router is trainable.
Therefore, the amortized communication overhead remains negligible.
Notably, compared with methods that transmit both LoRA matrices (i.e., $\mathcal{A}$ and $\mathcal{B}$), our design is more communication-efficient because only $\mathcal{B}$ is aggregated while $\mathcal{A}$ is kept fixed.

\begin{figure}[t]
    \centering
    \includegraphics[width=.83\linewidth]{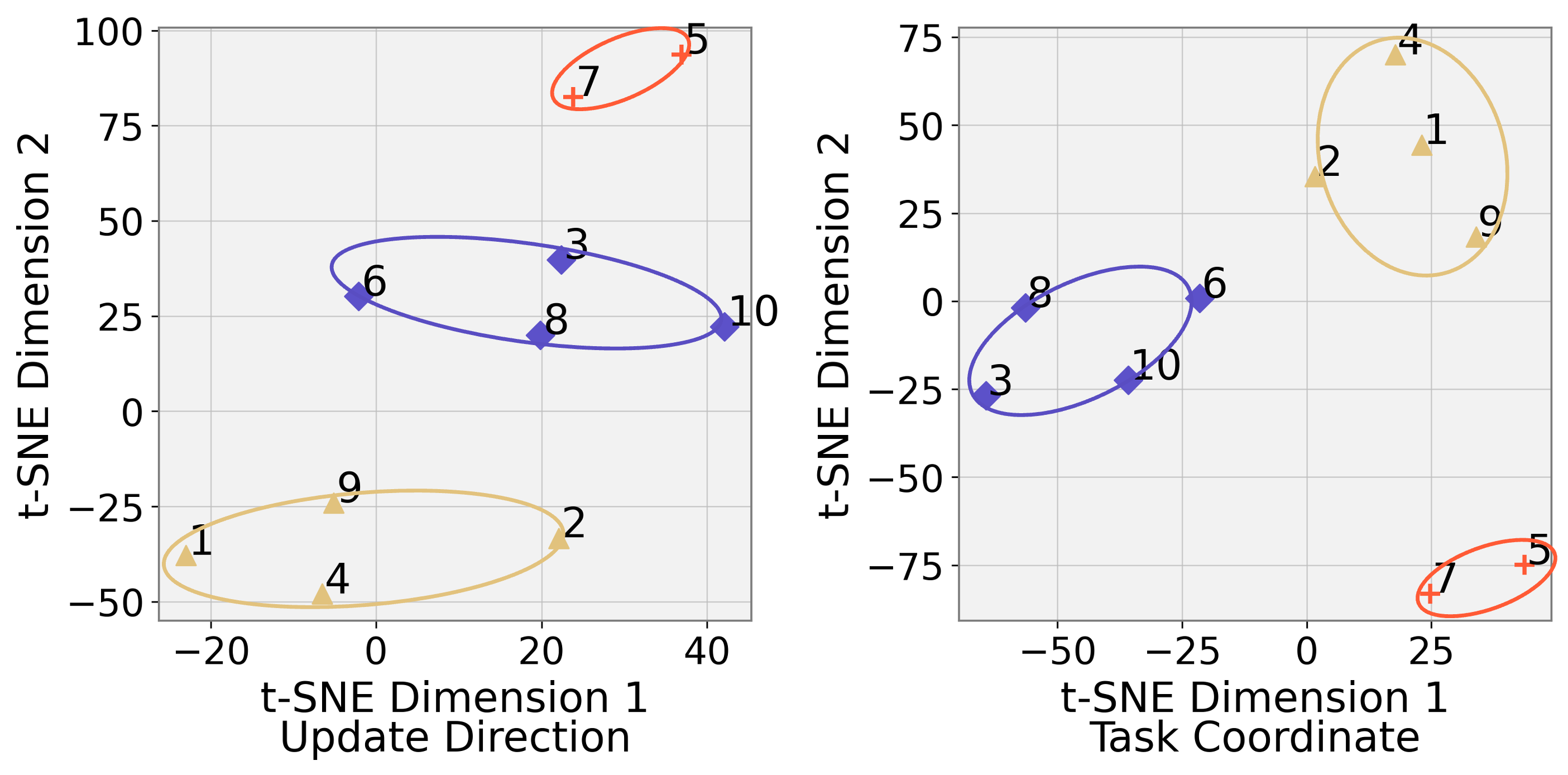}
    \caption{Task Structure vs. Update Directions (dolly)
    }
    \label{fig:pca}
\end{figure}
\begin{figure}[t]
    \centering
    \includegraphics[width=.7\linewidth]{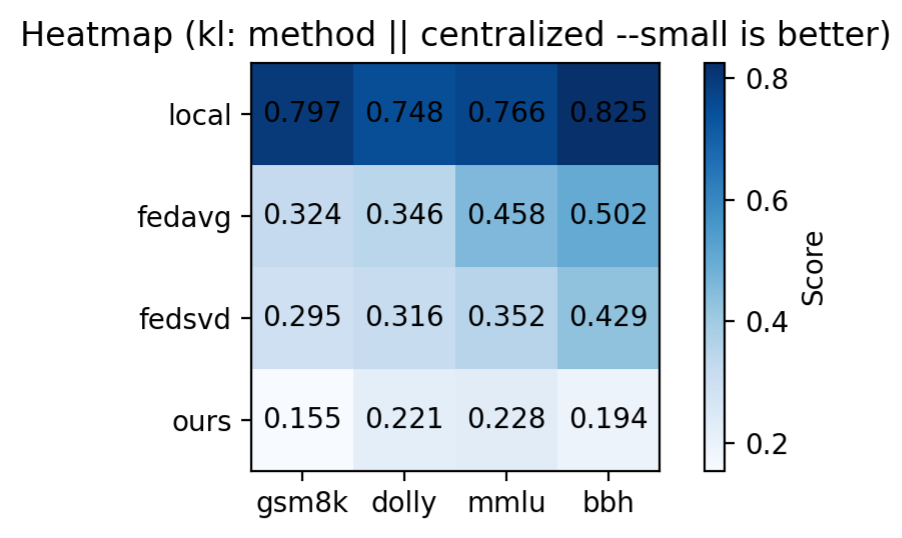}
    \caption{
    KL-divergence heatmap 
    }
    \label{fig:kl_heatmap}
\end{figure}
\begin{table*}[t]
\centering
\caption{Performance comparison on reasoning and knowledge benchmarks under different task heterogeneity levels. 
All reported results in the following tables are averaged over three random seeds, with standard deviation below 0.5\%.}
\resizebox{1\linewidth}{!}{
\renewcommand{\arraystretch}{1.1}
\begin{tabular}{l|ccc|ccc|ccc|ccc}
\toprule
Method      & \multicolumn{3}{c}{GSM8K (EM)} & \multicolumn{3}{c}{MMLU (ACC)} & \multicolumn{3}{c}{BBH (EM)} & \multicolumn{3}{c}{AVG.} \\
\midrule
$\alpha$    & 0.1 & 0.5 & 0.9& 0.1 & 0.5 & 0.9& 0.1 & 0.5 & 0.9& 0.1 & 0.5 & 0.9\\
\midrule
Base Model  
& 61.50\% & 61.50\% & 61.50\% 
& 62.50\% & 62.50\% & 62.50\% 
& 36.46\% & 36.46\% & 36.46\% 
& 53.49\% & 53.49\% & 53.49\% \\

Centralized 
& 68.00\% & 68.00\% & 68.00\% 
& 70.17\% & 70.17\% & 70.17\% 
& 42.57\% & 42.57\% & 42.57\% 
& 60.25\% & 60.25\% & 60.25\% \\

Local       
& 54.23\% & 62.01\% & 64.50\% 
& 57.85\% & 62.10\% & 62.90\% 
& 33.23\% & 36.60\% & 37.22\% 
& 48.44\% & 53.57\% & 54.87\% \\
\midrule
FedAvg ((AISTATS 2017)) \cite{mcmahan2017communication}  
& 52.16\% & 63.87\% & 66.59\%
& 56.94\% & 65.00\% & 64.30\% 
& 30.13\% & 36.72\% & 39.87\% 
& 46.41\% & 55.20\% & 56.92\% \\

FedProx  (MLSys 2020) \cite{li2020federated}
& 55.38\% & 63.95\% & 66.31\% 
& 58.26\% & 63.40\% & 64.00\% 
& 31.76\% & 37.36\% & 39.38\% 
& 48.47\% & 54.90\% & 56.56\% \\

SCAFFOLD (ICML 2020) \cite{karimireddy2020scaffold}
& 48.81\% & 52.86\% & 54.28\% 
& 54.77\% & 56.43\% & 57.32\% 
& 28.58\% & 30.52\% & 32.51\% 
& 44.05\% & 46.60\% & 48.04\% \\

FFA-LoRA (ICLR 2024)  \cite{sun2024improving}
& 62.29\% & 64.12\% & 65.03\% 
& 61.43\% & 62.87\% & 63.41\% 
& \cellcolor{second}36.46\% & 37.17\% & 38.08\% 
& \cellcolor{second}53.39\% & 54.72\% & 55.51\% \\

Fedex-LoRA  (ACL 2025) \cite{singhal2025fedex}
& 63.25\% & 63.01\% & 61.00\% 
& 61.68\% & 62.75\% & 63.57\% 
& 33.67\% & 35.00\% & \cellcolor{second}41.35\% 
& 52.87\% & 53.59\% & 55.31\% \\

FedSVD  (NeurIPS 2025) \cite{yan2025feddial}
& \cellcolor{second}63.11\% & \cellcolor{second}64.62\% & \cellcolor{second}66.71\%
& \cellcolor{second}62.06\% & \cellcolor{second}66.22\% & \cellcolor{second}65.86\% 
& 33.52\% & \cellcolor{second}38.21\% & 39.52\% 
& 52.90\% & \cellcolor{second}56.35\% & \cellcolor{second}57.36\% \\
\midrule
\textbf{FedTAR (ours)} 
& \cellcolor{best}64.37\% & \cellcolor{best}66.34\% & \cellcolor{best}67.43\% 
& \cellcolor{best}64.81\% & \cellcolor{best}68.95\% & \cellcolor{best}69.20\% 
& \cellcolor{best}37.98\% & \cellcolor{best}41.76\% & \cellcolor{best}41.95\% 
& \cellcolor{best}55.72\% & \cellcolor{best}59.02\% & \cellcolor{best}59.53\% \\
\bottomrule
\end{tabular}
}
\label{tab:main1}
\end{table*}

\begin{figure*}[h]
\centering
\includegraphics[width=0.6\linewidth]{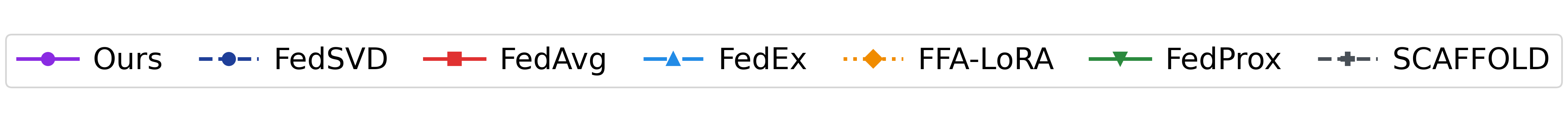}
\vspace{1em}
\begin{subfigure}[t]{0.46\textwidth}
    \centering
    \includegraphics[width=\linewidth]{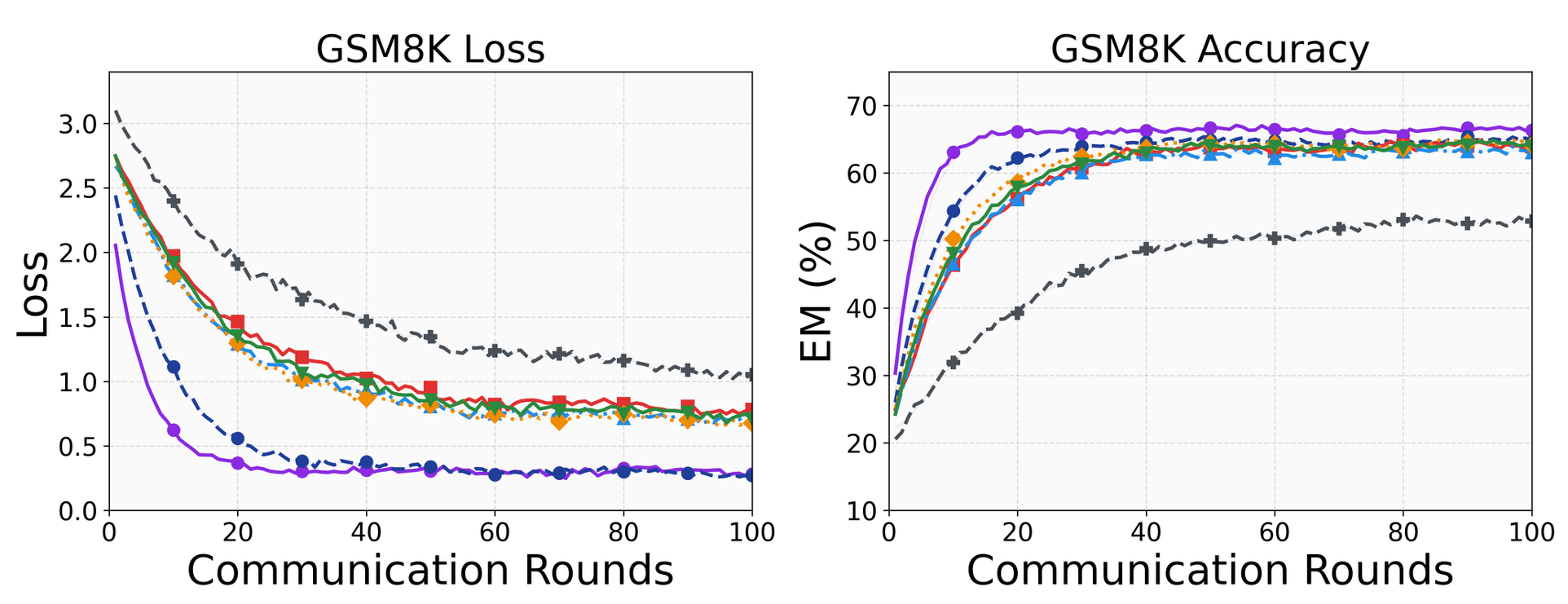}
    \caption{GSM8K}
\end{subfigure}
\begin{subfigure}[t]{0.46\textwidth}
    \centering
    \includegraphics[width=\linewidth]{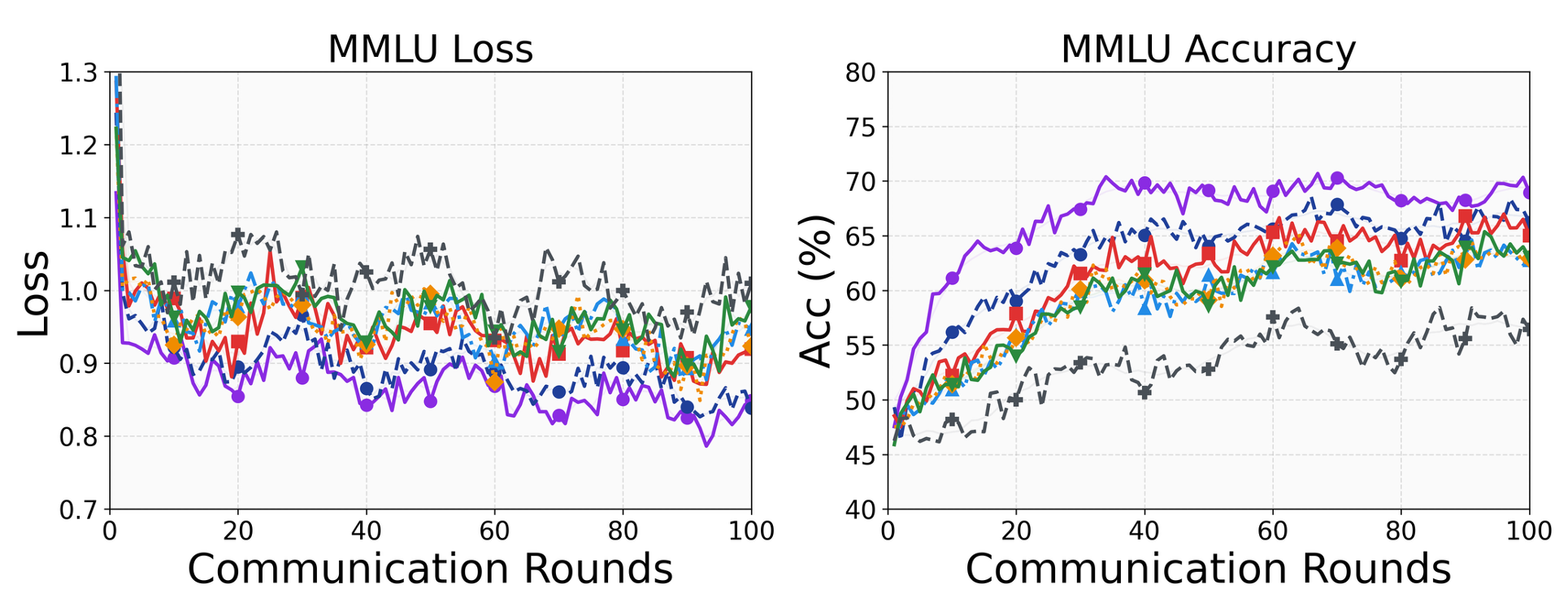}
    \caption{MMLU}
\end{subfigure}
\vspace{1em}
\begin{subfigure}[t]{0.46\textwidth}
    \centering
    \includegraphics[width=\linewidth]{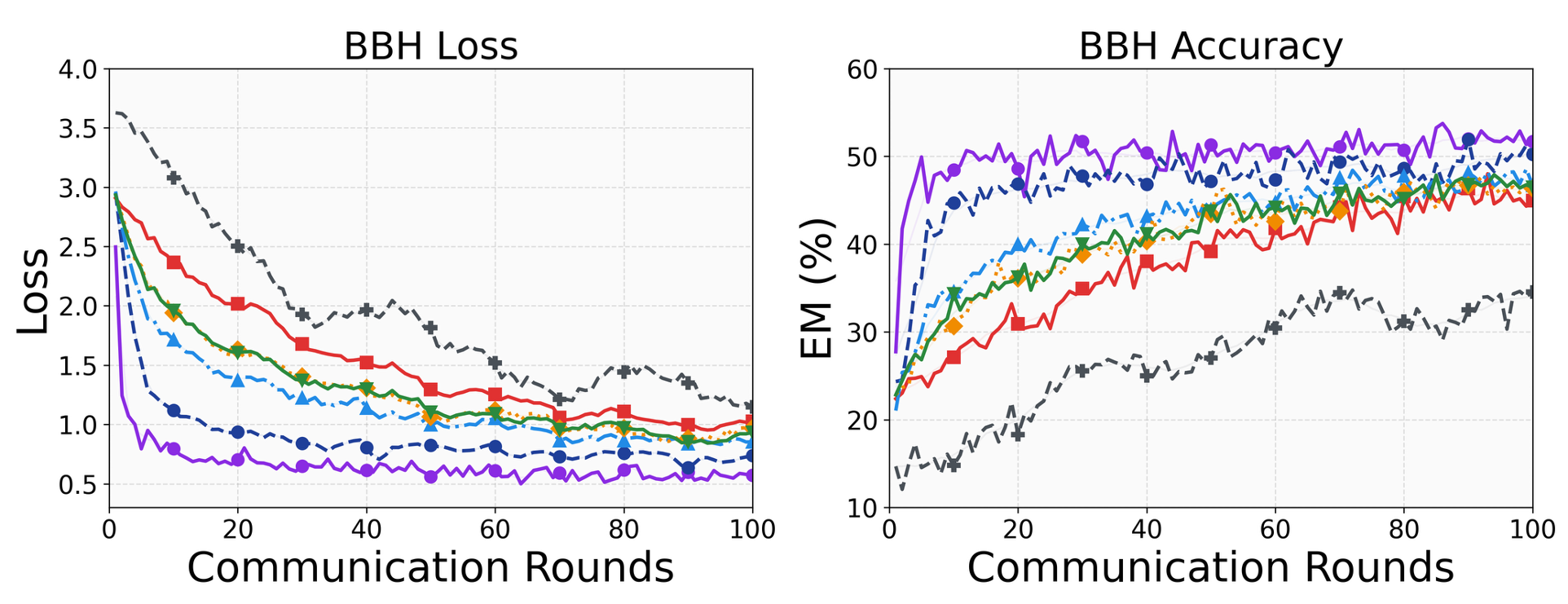}
    \caption{BBH}
\end{subfigure}
\begin{subfigure}[t]{0.46\textwidth}
    \centering
    \includegraphics[width=\linewidth]{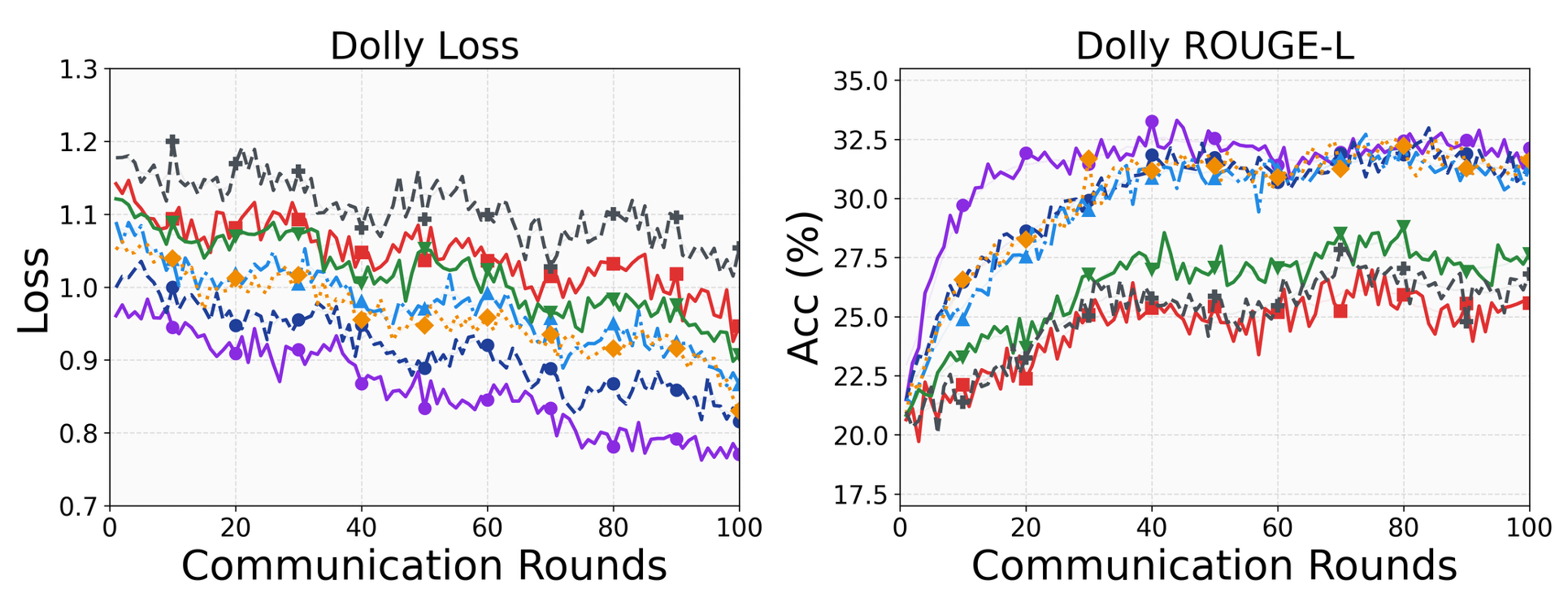}
    \caption{Dolly}
\end{subfigure}
\caption{
Training dynamics under federated settings. 
We report loss (left) and accuracy (right) across communication rounds on four datasets.
Our method achieves faster convergence and consistently better performance.
}
\label{fig:training_curves}
\end{figure*}
\begin{table}[]
\centering
\caption{Results on dolly (ROUGE-L).}
\renewcommand{\arraystretch}{1}
\begin{tabular}{@{}llll@{}}
\toprule
Method      & \multicolumn{3}{c}{Dolly-15K}  \\ \midrule
$\alpha$    & 0.1     & 0.5    & 0.9\\
\midrule
Base Model  
& 24.54\% & 24.54\% & 24.54\% 
\\

Centralized 
& 34.17\% & 34.17\% & 34.17\% \\

Local       
& 25.89\% & 27.03\% & 33.23\% 
\\
\midrule
FedAvg      
& 23.56\% & 25.57\% & \cellcolor{second}34.05\% 
\\

FedProx     
& 24.45\% & 27.65\% & 33.17\% 
\\

SCAFFOLD    
& 23.17\% & 26.80\% & 26.40\% 
\\

FFA-LoRA    
& 30.02\% & \cellcolor{second}31.60\% & 33.61\% 
\\

Fedex-LoRA  
& 30.58\% & 31.36\% & 33.00\% 
\\

FedSVD      
& \cellcolor{second}31.21\% & 31.53\% & 32.90\% 
\\
\midrule
\textbf{FedTAR (ours)}        
& \cellcolor{best}31.88\% & \cellcolor{best}32.12\% & \cellcolor{best}34.13\% \\

\bottomrule
\end{tabular}
\label{tab:main2}
\end{table}
\section{Experiments}
\subsection{Experimental Setup}
\paragraph{Datasets and Evaluations} We primarily evaluate the proposed method on a popular MoE-based sparse LLM: Qwen1.5-MoE-A2.7B \cite{qwen1_5_moe_a27b}. 
Qwen1.5-MoE-A2.7B has 14.3 billion parameters in total and 2.7 billion activated parameters during runtime with 64 fine-grained experts in each layer.
As a highly competitive open-source MoE model with fewer than 3B activated parameters,
Qwen1.5-MoE-A2.7B is particularly suitable for federated scenarios involving multiple resource-constrained devices.
Four benchmark tasks are used for evaluation: GSM8K (math reasoning) \cite{cobbe2021training}, MMLU (knowledge understanding) \cite{hendrycks2021measuring}, BBH (complex reasoning) \cite{kazemi2025big}, and Dolly-15K (instruction following) \cite{conover2023free}.
We evaluate model performance using task-specific metrics. 
For GSM8K and BBH, Exact Match (EM) accuracy is reported. 
For MMLU, we use classification accuracy (ACC). 
For Dolly-15K, we use ROUGE-L to measure quality.

\begin{table}[t]
\centering
\small
\setlength{\tabcolsep}{6pt}
\caption{Ablation study of the proposed method across four datasets ($\alpha=0.5$).}
\label{tab:ablation}
\renewcommand{\arraystretch}{1}
\resizebox{0.49\textwidth}{!}{
\begin{tabular}{lccccc}
\toprule
\textbf{Variant} & \textbf{GSM8K} & \textbf{Dolly} & \textbf{MMLU} & \textbf{BBH} & \textbf{Avg.} \\
\midrule
Full model                  & 66.34\%& 32.12\%& 68.95\%& 41.76\%& 52.29\%\\
w/o SVD& 64.17\%& 26.56\%& 66.24\%& 37.31\%& 48.57\%\\
w/o clustering              & 63.65\%& 27.01\%& 67.87\%& 39.14\%& 49.42\%\\
w/o reconstruction          & 64.83\%& 27.75\%& 67.43\%& 39.77\%& 49.95\%\\
mean-only router    & 65.26\%& 31.24\%& 68.32\%& 41.34\%& 51.54\%\\
 coact-only router   & 65.67\%& 31.58\%& 68.51\%& 41.56\%&51.83\%\\
w/o router (FedAvg)& 63.87\%& 25.57\%& 65.00\%& 36.72\%& 47.79\%\\
\bottomrule
\end{tabular}
}
\vspace{-0.6cm}
\end{table}

\paragraph{Comparison Methods}
We compare FedTAR with several FL baseline methods and state-of-the-art federated LLM fine-tuning methods.
\textbf{1. FedAvg \cite{mcmahan2017communication}:} Baseline FL method. Both LoRA matrix $\mathcal{A}$ and $\mathcal{B}$ are trained locally and averaged on server.
\textbf{2. FedProx \cite{li2020federated}:} The standard federated learning baseline. An extension of FedAvg with a proximal regularization term, which mitigates client drift under heterogeneous data distributions.
\textbf{3. Scaffold \cite{karimireddy2020scaffold}:} A variance reduction method, correcting client drift by control variates.
\textbf{4. FFA-LoRA \cite{sun2024improving}:} It fixes matrix $\mathcal{A}$ and only update $\mathcal{B}$ for decreasing the server aggregation bias.
\textbf{5. FedEx-LoRA \cite{singhal2025fedex}:}  After training $\mathcal{A}$ and $\mathcal{B}$, the redefined residual is added to pre-trained matrix.
\textbf{6. FedSVD \cite{yan2025feddial}:} Only $\mathcal{B}$ is updated on server. Client refactorizes $\mathcal{BA}$ locally via SVD.
\textbf{FedTAR (ours):} 
Only the $\mathcal{B}$ is updated while keeping $\mathcal{A}$ fixed. 
We leverage routing features to reconstruct task-aware updates.

\paragraph{Implementation Details}
We adopt Alpaca-GPT-4 dataset for training and evaluation on GSM8K, BBH and MMLU, following general instruction-tuning protocol.
To simulate data heterogeneity in FL, 
the training data are partitioned across clients by Dirichlet distribution.
For Dolly-15K (instruction following), we follow the setting of \cite{qin2025federated}.
Specifically, Dolly-15K contains 15015 data samples corresponding to 8 tasks.
1188 samples from one task are reserved for evaluation,
while the remaining date are used for training.
The training data are partitioned across clients via a Dirichlet distribution based on the task category
This setup introduces a more challenging form of task heterogeneity.

The training data are partitioned among 10 clients using a Dirichlet distribution with $\alpha \in {0.1, 0.5, 0.9}$ to simulate different levels of non-IID data distributions.
A smaller $\alpha$ indicates stronger heterogeneity.
LoRA is used with rank $r=8$, scaling factor 16, and dropout rate 0.05.
Each round takes approximately 2 to 3 minutes.
The learning rate is set to $1\times10^{-5}$, and the maximum input sequence length is 256 tokens.
Each client uses a local batch size of 16.
All clients are trained for 100 communication rounds.
In each round, each client locally updates the parameters using vanilla SGD for $\tau=5$ steps, with the gradient clipping norm set to 2.
We incorporate differential privacy (DP) into federated training to protect transmission privacy with $\epsilon = 6$.
The standard deviation is below 0.5\% for all reported results.
All experiments are conducted on a single NVIDIA RTX Ada 6000 Pro GPU.

\subsection{Insight Validation}
We first verify our insight that task space and update space exhibit a corresponding relationship.
Fig. \ref{fig:pca} visualizes the client distributions in both update directions and task coordinates using t-SNE on Dolly-15K with 10 clients and $\alpha=0.5$.
The clustering results in the two spaces exhibit a high-level consistency.
For example, clients 5 and 7 are grouped into the same cluster in both spaces, indicating that their update behaviors are aligned with their underlying task representations.
In addition, different clusters in the update space show distinct convergence directions.
For instance, the yellow and red clusters are clearly separated, revealing potential update conflicts under diverse tasks.
Furthermore, Fig. \ref{fig:kl_heatmap} shows the KL divergence \cite{li2026expert} between expert usage distributions of different FL methods and that of the centralized training model, where lower values indicate better consistency.
FedTAR achieves the lowest KL divergence, outperforming local training, FedAvg, and FedSVD.
This demonstrates that task-aware aggregation better preserves task structure and reduces update conflicts, making the aggregated updates more consistent with centralized training.

\subsection{Main Results}

\paragraph{Performance Evaluation}
Tables \ref{tab:main1} and \ref{tab:main2} present the performance of FedTAR and comparison methods under varying levels of data heterogeneity.
The best and second-best results are highlighted in green and yellow, respectively.
All baselines suffer performance degradation as data heterogeneity increases. 
For example, on GSM8K, FedAvg drops from 0.6639 ($\alpha=0.9$) to 0.5126 ($\alpha=0.1$), while FedSVD decreases from 0.6671 to 0.6311. 
In particular, these baseline methods lack explicit task awareness, 
making them prone to conflicting updates. 
In contrast, FedTAR consistently achieves \textit{SOTA} across all settings. 
For instance, under $\alpha=0.1$, FedTAR reaches 0.6437, outperforming FedAvg (0.5126) and FedSVD (0.6311), and further improves to 0.6743 at $\alpha=0.9$. 
Similar trends are observed on other datasets, 
with more pronounced gains on Dolly-15K where task heterogeneity is stronger.
Under $\alpha=0.1$, FedTAR achieves 0.3188, outperforming FedAvg (0.2356) and SCAFFOLD (0.2317), and remaining competitive with FedSVD (0.3121).
\paragraph{Training Curves}
FedTAR improves communication efficiency by achieving higher performance within fewer communication rounds through faster convergence.
As shown in Fig. \ref{fig:training_curves}, on GSM8K and BBH, FedTAR reaches its best performance within 10 rounds, while the second-best method, FedSVD, requires around 30 rounds.
On MMLU and Dolly-15K, despite more fluctuating loss curves, FedTAR consistently maintains superior performance.

Overall, FedTAR achieves both superior performance and improved communication efficiency under different heterogeneous settings.

\subsection{Ablation Study}
To quantify the effectiveness of each component in our design, we conduct ablation studies by comparing the full model with several variants:
(1) w/o SVD: removing the SVD decomposition and directly aggregating full updates and routing features on the server;
(2) w/o clustering: applying reconstruction to the FedAvg result without task-based clustering;
(3) w/o reconstruction: performing task-based clustering and aggregation without the reconstruction step;
(4) mean-only router: using only the average expert usage as the routing feature;
(5) coact-only router: using only the average expert co-activation as the routing feature;
(6) w/o router: removing routing features entirely, which degenerates to FedAvg.

As shown in Table \ref{tab:ablation}, removing any component leads to performance degradation.
Among all variants, w/o router results in the most significant drop, with the average performance decreasing from 52.29\% to 47.79\%.
Removing SVD also leads to a notable decrease, from 52.29\% to 48.57\%.
This shows that SVD effectively extracts structured update directions and task coordinates.
Without SVD, the model tends to learn entangled signals, limiting the update--task alignment mechanism.
Meanwhile, w/o clustering and w/o reconstruction further degrade performance, indicating that both task grouping and task-aligned reconstruction are essential for mitigating update conflicts.
Furthermore, using partial routing features, such as only the mean expert usage (51.54\%) or only expert co-activation (51.83\%), is inferior to the full model (52.29\%), indicating that a richer task representation is needed to fully capture task heterogeneity.

\subsection{Parameters}
\paragraph{Impact of client number in FedTAR}
\begin{figure}[t]
    \centering
    \includegraphics[width=.96\linewidth]{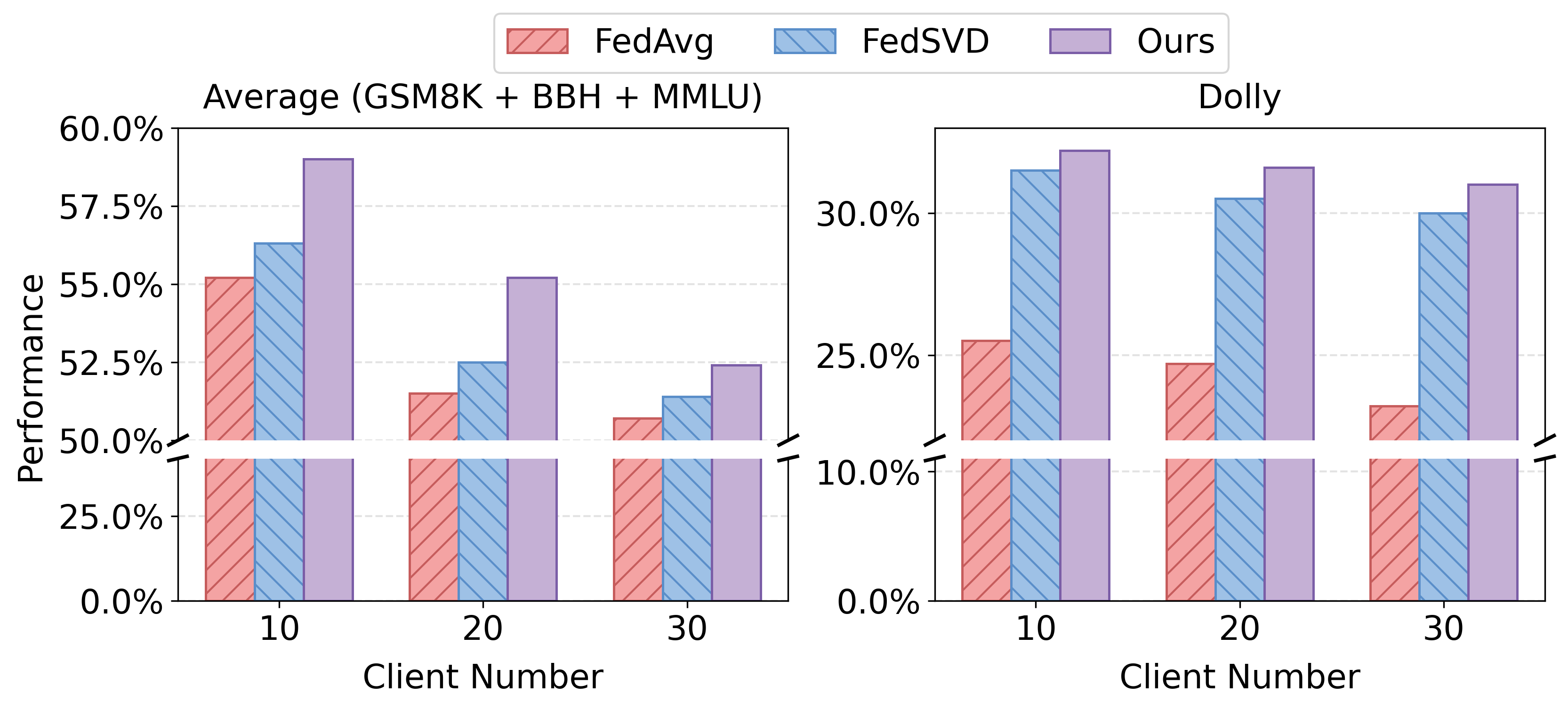}
    \caption{Impact of number of clients in FedTAR ($\alpha=0.5$).
    }
    \label{fig:num_clients}
\end{figure}
We first investigate the impact of the number of clients.
As illustrated in Fig. \ref{fig:num_clients}, FedTAR consistently outperforms the baseline methods, FedAvg and FedSVD, regardless of the number of clients.
For example, on the GSM8K+BBH+MMLU tasks, FedTAR achieves approximately 0.59, 0.55, and 0.53 with 10, 20, and 30 clients, respectively, while FedAvg drops from approximately 0.55 to 0.51.
A similar trend is observed on Dolly: as the number of clients increases, FedTAR maintains the best performance, decreasing only slightly from approximately 0.32 to 0.31.
These results indicate that although performance slightly drops as the number of clients increases, FedTAR remains more robust than the baseline methods.

\paragraph{Impact of LoRA rank in FedTAR}
\begin{figure}[t]
    \centering
    \includegraphics[width=1\linewidth]{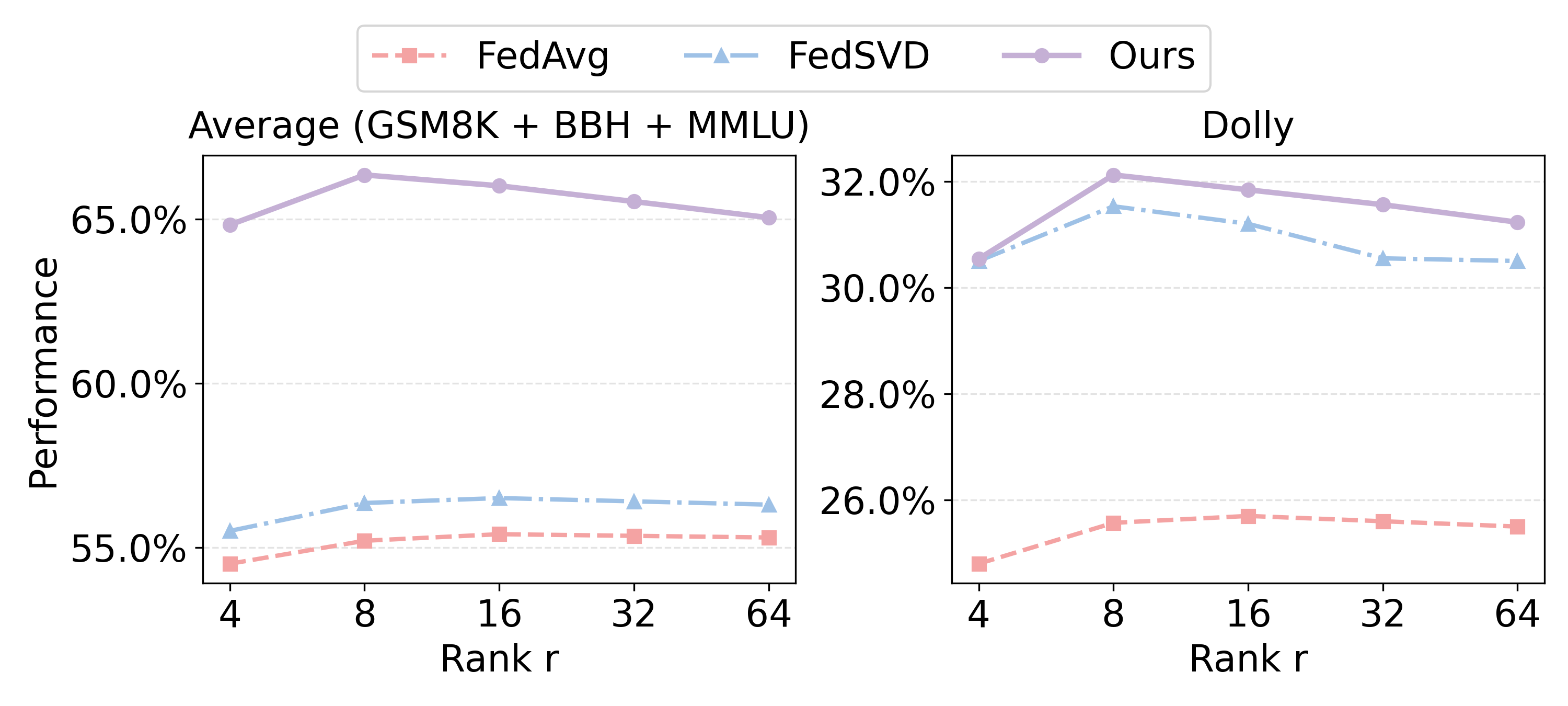}
    \caption{Impact of number of rank in FedTAR ($\alpha=0.5$).
    }
    \label{fig:rank}
\end{figure}
We further analyze the effect of LoRA rank in Fig. \ref{fig:rank}.
We tested the model performance for LoRA ranks ranging from 4 to 64.
As $r$ increased from 4 to 8, FedTAR’s performance improved from approximately 0.65 to 0.66 on the GSM8K+BBH+MMLU dataset and from 0.31 to 0.32 on the Dolly dataset; performance then plateaued. In contrast, baseline methods such as FedAvg show only marginal improvements (e.g., approximately 0.55 to 0.56). This suggests that a moderate rank is sufficient for FedTAR to capture the fundamental update structure while avoiding unnecessary parameter overhead.

\paragraph{Impact of cluser number in FedTAR}
\begin{figure}[t]
    \centering
    \includegraphics[width=1\linewidth]{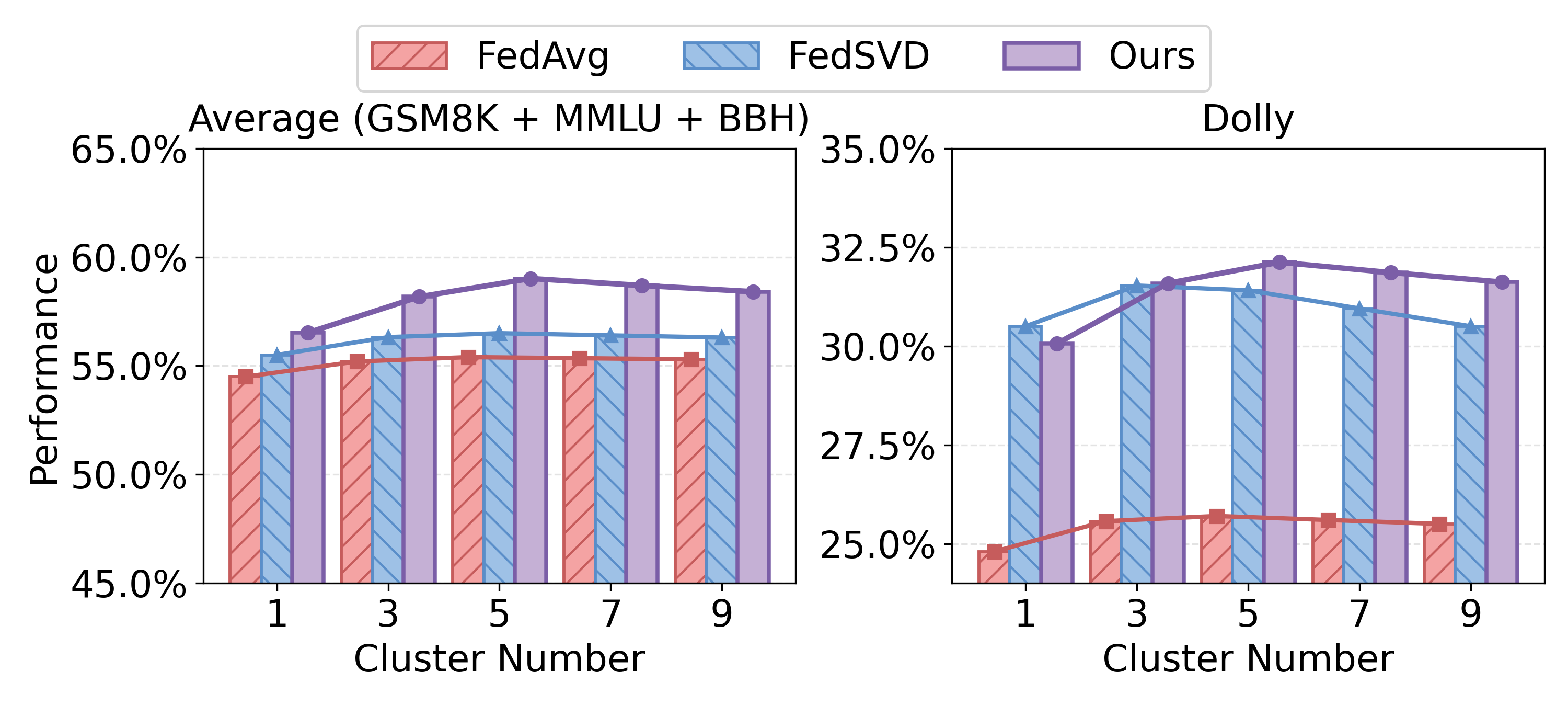}
    \caption{Impact of number of cluster in FedTAR ($\alpha=0.5$).
    }
    \label{fig:cluster}
\end{figure}
Finally, we examine the impact of the number of clusters in Fig. \ref{fig:cluster}.
Increasing the number of clusters from 1 to 5 improves performance from approximately 0.57 to 0.59 on the GSM8K+MMLU+BBH dataset, and from 0.30 to 0.32 on the Dolly dataset.
However, further increasing the number of clusters leads to a slight performance decline or a plateau.
This indicates a trade-off between capturing finer-grained task structures and maintaining sufficient data within each cluster.
FedTAR maintains the best performance across all settings.

\section{Conclusion}
In this paper, we propose FedTAR, a task-aware federated fine-tuning method for MoE-based LLMs.
FedTAR leverages routing features to establish a mapping between task coordinates and update directions.
Based on the task coordinates, the server aggregates local updates through clustering and reconstructs the aggregated result using the learned mapping.
In this way, FedTAR preserves expert specialization and mitigates conflicting updates on shared experts.
Experimental results on four tasks demonstrate that FedTAR achieves state-of-the-art performance, highlighting the importance of task-aware MoE aggregation for heterogeneous federated fine-tuning.

\section{Acknowledge}
This work was supported by the National Science Foundation of U.S. (2416872, 2315596, 2244219, 2146497, 2551417, 2548961).






\bibliography{ref}{}
\bibliographystyle{IEEEtran}

\appendix

\subsection{Effectiveness of Task-Aware Alignment}

We provide a principled justification for why routing-based task representations
can guide update aggregation.

\paragraph{Routing captures task structure.}
In MoE models, the output for an input $x$ is
\begin{equation}
f(x) = \sum_{k=1}^K g_k(x) E_k(x),
\end{equation}
where $g_k(x)$ is the routing probability over experts.
For a client dataset $\mathcal{D}_i$, define the average routing pattern:
\begin{equation}
\mu_i = \mathbb{E}_{x \sim \mathcal{D}_i}[g(x)],
\end{equation}
and co-activation matrix
\begin{equation}
C_i = \mathbb{E}[g(x) g(x)^\top].
\end{equation}

These statistics summarize how inputs are distributed across experts.
Since different tasks typically activate different subsets of experts,
$(\mu_i, C_i)$ encode the task preference of client $i$.
In particular, clients with similar data distributions induce similar routing patterns.

\paragraph{From routing to updates.}
Local updates are driven by gradients:
\begin{equation}
\Delta w_i = -\eta \, \mathbb{E}_{x \sim \mathcal{D}_i} \nabla \ell(f(x)).
\end{equation}
Since $f(x)$ depends on routing weights $g(x)$,
the gradient depends on how tokens are assigned to experts.
Under standard smoothness assumptions on the loss $\ell$
and expert networks $E_k$, the mapping from routing behavior to updates is continuous.

We therefore model this relationship as
\begin{equation}
\Delta w_i = \Phi(z_i) + \xi_i,
\end{equation}
where $z_i$ is the routing-based representation,
$\|\xi_i\| \le \sigma$, and $\Phi$ is $L$-Lipschitz:
\begin{equation}
\|\Phi(z_i) - \Phi(z_j)\|
\le
L \|z_i - z_j\|.
\end{equation}

\paragraph{Implication for aggregation.}
For any clients $i,j$, we obtain
\begin{equation}
\|\Delta w_i - \Delta w_j\|
\le
L \|z_i - z_j\| + 2\sigma.
\end{equation}
Thus, proximity in the task space implies proximity in update directions.
This provides a theoretical basis for clustering clients using routing features
and performing task-aware aggregation.

\paragraph{Low-rank structure of updates.}
Empirically, client updates often lie in a low-dimensional subspace due to shared model structure.
We model this as
\begin{equation}
\Delta w_i = U a_i + \epsilon_i,
\end{equation}
where $U \in \mathbb{R}^{d \times m}$ is an orthonormal basis
(e.g., obtained via SVD over client updates).

Given aggregated update
\begin{equation}
\Delta w_{\mathrm{agg}} = \sum_i \alpha_i \Delta w_i,
\end{equation}
define
\begin{equation}
\Delta w_{\mathrm{TA}} = U U^\top \Delta w_{\mathrm{agg}}.
\end{equation}

Let
\begin{equation}
\Delta w^\star = \sum_i \alpha_i U a_i.
\end{equation}

Then
\begin{equation}
\|\Delta w_{\mathrm{TA}} - \Delta w^\star\|
\le
\|\Delta w_{\mathrm{agg}} - \Delta w^\star\|.
\end{equation}

Moreover, the error admits the decomposition
\begin{equation}
\|\Delta w_{\mathrm{agg}} - \Delta w^\star\|^2
=
\|\Delta w_{\mathrm{TA}} - \Delta w^\star\|^2
+
\|(I - U U^\top)\sum_i \alpha_i \epsilon_i\|^2.
\end{equation}

This shows that task-aware projection removes the component orthogonal to the shared subspace,
which corresponds to conflicting or task-inconsistent update directions.

\subsection{Least-Squares Alignment and Projection}

We detail the computation of the task--update mapping used in the main method.

Let
\[
Z = [z_1, \dots, z_N] \in \mathbb{R}^{r \times N}, \quad
\Delta = [\Delta w_1, \dots, \Delta w_N] \in \mathbb{R}^{d \times N}.
\]

We aim to approximate the (unknown) mapping $\Phi$ from task space to update space.
To this end, we solve the ridge regression problem:
\begin{equation}
\mathbf{M}
=
\arg\min_{\mathbf{M}}
\|\Delta - \mathbf{M} Z\|_F^2 + \lambda \|\mathbf{M}\|_F^2.
\end{equation}

Expanding the objective,
\[
\|\Delta - \mathbf{M} Z\|_F^2
=
\mathrm{Tr}((\Delta - \mathbf{M}Z)(\Delta - \mathbf{M}Z)^\top),
\]
and taking derivative with respect to $\mathbf{M}$ yields the normal equation:
\begin{equation}
\mathbf{M} Z Z^\top + \lambda \mathbf{M} = \Delta Z^\top.
\end{equation}

Solving for $\mathbf{M}$ gives
\begin{equation}
\mathbf{M}
=
\Delta Z^\top (Z Z^\top + \lambda I)^{-1}.
\end{equation}

Thus, $\mathbf{M} z_i$ provides the best linear approximation (under squared loss)
to the update induced by task $z_i$. The column space of $\mathbf{M}$ therefore
defines a task-aligned update subspace that captures the dominant variation
of client updates conditioned on tasks.

Given an aggregated update $\Delta'$, we project it onto this subspace by solving
\begin{equation}
\min_a \|\Delta' - \mathbf{M} a\|_2^2 + \lambda \|a\|_2^2,
\end{equation}
whose solution is
\begin{equation}
a^\star =
(\mathbf{M}^\top \mathbf{M} + \lambda I)^{-1}
\mathbf{M}^\top \Delta'.
\end{equation}

Substituting back yields the projected update:
\begin{equation}
\Delta_{\mathrm{proj}}
=
\mathbf{M}
(\mathbf{M}^\top \mathbf{M} + \lambda I)^{-1}
\mathbf{M}^\top \Delta'.
\end{equation}

This projection restricts the aggregated update to directions that can be expressed
as combinations of task-conditioned update bases, thereby filtering out components
that are inconsistent with any learned task structure.
\end{document}